\documentclass[lettersize,journal]{IEEEtran}
\usepackage{amsmath,amsfonts}
\usepackage{algorithmic}
\usepackage{algorithm}
\usepackage{array}
\usepackage[caption=false,font=normalsize,labelfont=sf,textfont=sf]{subfig}
\usepackage{textcomp}
\usepackage{stfloats}
\usepackage{url}
\usepackage{verbatim}
\usepackage{graphicx}
\usepackage{cite}
\usepackage{multirow}
\begin{document}

\title{Socially reactive navigation models for mobile robots in dynamic environments}

\author{Ricarte de Sousa Ribeiro, Plinio Moreno \\ ricarte.ribeiro@tecnico.ulisboa.pt, plinio@isr.tecnico.ulisboa.pt \\ Instituto Superior T\'ecnico, Lisboa, Portugal}

% The paper headers
%\markboth{Journal of \LaTeX\ Class Files,~Vol.~14, No.~8, August~2021}%
%{Shell \MakeLowercase{\textit{et al.}}: A Sample Article Using IEEEtran.cls for IEEE Journals}

%\IEEEpubid{0000--0000/00\$00.00~\copyright~2021 IEEE}
% Remember, if you use this you must call \IEEEpubidadjcol in the second
% column for its text to clear the IEEEpubid mark.

\maketitle

\begin{abstract}
The objective of this work is to expand upon previous works, considering socially acceptable behaviours within robot navigation and interaction, and allow a robot to closely approach static and dynamic individuals or groups. The space models developed in this dissertation are adaptive, that is, capable of changing over time to accommodate the changing circumstances often existent within a social environment. The space model's parameters' adaptation occurs with the end goal of enabling a close interaction between humans and robots and is thus capable of taking into account not only the arrangement of the groups, but also the basic characteristics of the robot itself. This work also further develops a preexisting approach pose estimation algorithm in order to better guarantee the safety and comfort of the humans involved in the interaction, by taking into account basic human sensibilities. The algorithms are integrated into ROS's navigation system through the use of the $costmap2d$ and the $move\_base$ packages. The space model adaptation is tested via comparative evaluation against previous algorithms through the use of datasets. The entire navigation system is then evaluated through both simulations (static and dynamic) and real life situations (static). These experiments demonstrate that the developed space model and approach pose estimation algorithms are capable of enabling a robot to closely approach individual humans and groups, while maintaining considerations for their comfort and sensibilities.
\end{abstract}


%\maketitle

\section{Introduction}
\label{sec:intro}

With the continuous advancements in robotics, robots have been steadily moving away from purely industrial and controlled environments to environments populated by people of all walks of life. Many such robots already exist within public spaces and even people's own homes \cite{lee2021service}. Few of these robots, though, are capable of properly accommodating a human's sensibilities. It is thus necessary to study the nature of social interaction and how to best adapt it to robotics.

The greater motivation for this dissertation, within the aforementioned context, is the importance of a robot within a social environment being capable of proactively interacting with an individual or a group of individuals in a socially acceptable manner. This requires that the personal and group spaces are modelled in a way to facilitate the robot's navigation for interaction purposes and to properly calculate a position for the robot to execute an approach and initiate this interaction, both in static and dynamic environments.

Most state-of-the-art methods utilize fixed parameters within their social navigation systems, which leads to a degree of inflexibility. The methods that do possess adaptive qualities are often restricted to static environments or utilize adaptations oriented towards human avoidance. In order to enable more comfortable human-robot interactions, several factors must be taken into account. First, humans have flexible considerations of their own personal or group spaces that are influenced by the situations they are inserted into. It is necessary for space models to be adaptive in order to properly address this factor. Secondly, a human within a social space is often in motion. Preparing a robot to take into account the dynamic nature of people is essential to guaranteeing their safety and comfort. Lastly, robots are variable in nature, especially in a constantly changing field such as social robotics. It is thus necessary to be able to adapt existing considerations to the spatial characteristics of the robot itself. The work presented in this dissertation thus aims to be able to adapt the personal and group space models according to these aspects and enable proactive human-robot interaction. The main problem is the selection of a proper pose for interaction with individuals and groups, ensuring as much safety and comfort as possible.

Several modules are necessary for this work, such as the proper detection of humans and their pose (position and orientation) and of the groups they might form, in order to properly determine what parameters suit each unique situation. 

This work has the following contributions: (i) Flexible personal and group space models for human-robot interaction, that take into account dynamic considerations and the robot's characteristics, (ii) An adaptation to an approach pose estimator to ensure the safety of humans involved in social situations and the dynamic nature thereof, (iii) Changes to previously developed ROS packages to integrate these changes. These are: a package that provides ROS messages that contain group information \footnote{\url{https://github.com/Ricarte-Ribeiro/group_msgs}}, a package responsible for handling the calculation of the personal and group space parameters and their integration within a costmap\footnote{\url{https://github.com/Ricarte-Ribeiro/adaptive_social_layers}}, and a package that estimates an approach pose for a chosen individual or group \footnote{\url{https://github.com/Ricarte-Ribeiro/approach_group}}. 

The remainder of this document is organized as follows: Section \ref{sec: backg} presents the background and related works, Section \ref{sec: Methodology} describes the approach taken to the subject matter of the dissertation and presents preliminary results, Section \ref{sec:Experiments} describes how the algorithms are integrated into the navigation system and presents experimental results, and Section \ref{sec:concl} presents the conclusions of this work.

\section{Related Works}
\label{sec: backg}

\subsection{Background}

\subsubsection{Mapping}

The common representation of obstacles and free space is the occupancy grid method used to fill out the map. \cite{moravec1989sensor}\cite{matthies1988integration}. It divides the space into a grid of evenly spaced cells represented by binary random variables. The variables, depending on their values, represent whether a cell is occupied and thus non-traversable. At each cell, its cost is computed from the detected obstacles. To increase performance, Lu et. al. \cite{lu2014layered} introduced the multi-layered costmap, where each layer includes different information that can be merged in a custom manner by the user. This also allows for only areas of the costmap that suffered changes to be updated, removing the need to fully update the master costmap.

\subsubsection{Navigation}

Navigation occurs through the use of planners to calculate a path and translate it into velocity commands. There are two specific types of planners relevant to this work: global planners and local planners. Global planners take into account the entire breadth of the mapping information available to the robot and calculate a route through which it can reach its goal, but require a map. These algorithms are slower than their local counterparts. Local planners, on the other hand, can address any immediate problems the robot might face during its navigation, detected by its sensors, due to their smaller runtime.

\subsubsection{Proxemics}
\label{Proxemics}

Hall \cite{ehall1990}, who coined the term proxemics, defined it as "the interrelated observations and theories of humans' use of space as a specialized elaboration of culture", and studied how humans manage the space around them during interpersonal communication.

Personal Space: Hall defined personal space as concentric circles around a human and divided it into four parts:(i) Intimate Space ($x<0.45m$), (ii) Personal Space ($0.45<x<1.2m$), (iii) Social Space ($1.2<x<3.6m$) and (iv) Public Space ($x>3.6m$).

Several studies since have studied different forms of shaping the personal space. \cite{hayduk1981shape} determined that people give greater importance to their front, and thus personal space is larger in that direction, giving it an egg-shape. \cite{helbing1995social} proposed that personal space is defined as elliptical equipotential lines centered on the person and oriented in the direction of motion. \cite{gerin2008characteristics} theorized that personal space could be smaller on the person's dominant side. \cite{hayduk1994personal} determined that personal space is not a constant and can shift in response to differing circumstances. \cite{park2007multi} introduced the concept of a spatio-temporal personal space that can be adapted taking into account the person's velocity. This work's approach is justified in these last two works.

Group Space: During social interaction, multiple individuals can form a group which follows its own specific rules in terms of spatial interaction. This group can be defined through the knowledge of the people's positions and orientations and multiple forms of defining it have been studied. The most well known is Adam Kendon's Facing Formation (F-Formation) \cite{kendon_2009}. Kendon organizes an F-Formation into three different spaces, defined through concentric circles: o-space, the innermost space of the group, delimited by every member of the group and which no stranger may breach; p-space, the area of space after the o-space and where all members of a group are located; and r-space, the area surrounding o-space and p-space. It is the last area separating a group from the rest of the world and it is also where people must move to in order to leave or enter the group.

\subsection{Human perception of a robot's approach}

\cite{joosse2014cultural} determined that people prefer a robot that approaches a group at a distance around 80 to 100 $cm$ from the center. They also noted that people prefer that the robot does not breach the intimate space of any of the group members and that the comfort of the approach may depend on which individuals the robot inserts itself between when entering the group. \cite{walters2007robotic,woods2006methodological} studied the approach direction of a robot. They concluded that frontal approaches are preferred and rear approaches cause discomfort. \cite{ball2014group} study the comfortability of the approach direction of a robot in regard to a group of people. They concluded that the approach direction does not matter as long as the robot is within the field of view of a group member.

\subsection{Socially-Aware Robot Navigation - Literature Review}

\subsubsection{Space Models and approach pose estimation}

 \cite{amaoka2009personal,kirby2010social,vega2018flexible,franciscomelo} model personal space through the use of a merger of two 2D Gaussian functions, to represent the difference in the rear and frontal directions of the personal space. \cite{amaoka2009personal}'s model requires a lookup table that has to be adjusted a priori in order to properly function and the capability to differentiate the age and gender of humans. \cite{kirby2010social}'s model is dependent on velocity to determine orientation and thus remains a circle when the person is static. \cite{vega2018flexible,franciscomelo} utilize adaptive parameters in order to create a flexible space model. \cite{vega2018flexible}'s model can adapt to spatial context and human intention. \cite{franciscomelo} expands upon \cite{vega2018flexible} enabling the adaptation to work on group space too and further adapts the personal space of people in a group to ensure that there is no overlap between each individual's personal space. This last work is the primary inspiration for this dissertation.

 \cite{papadakis2014adaptive} utilizes a skew-normal probability density function to define a personal space that can adapt to the certainty the robot possesses of a human’s characteristics. It utilizes four shapes defined in Section \ref{Proxemics}.

\cite{truong2016dynamic} defines the concept of Dynamic Social Zone (DSZ), defined by two parts. The first of them, the Extended Personal Space, expands Hall’s idea of personal space by taking into account several new elements related to human sensibilities. The other part, the Social Interaction Space, covers the o-space of a group or the possibility of an interaction between a human and an object. This work also goes on to describe a method to determine approach poses by ensuring that they must be located within the field of view of the individuals and outside the DSZ. Later in \cite{truong2017approach}, the concept is expanded to take into account the status of the human (sitting, standing, etc).

 \cite{livramento2020natural} present a data-driven model to estimate an appropriate approaching pose. The model was trained utilizing a dataset consisting of real life situations. While the algorithm estimates approach poses with a good success rate, it is dependent on the size of the training set, and there is also the possibility of failure for a situation sufficiently different from any within it.

 \cite{gomez2014fast} utilizes the Fast Marching Square method ($FM^2$) \cite{valerogomez2013fast} of path planning with an adaptation for social environments. This method lacks experimental validation with humans. There is also the issue that the approach method chosen does not take into account the directionality of the approach or more than the o-space of a group.

 \cite{rios2011understanding} extends the preexisting Risk-RRT method \cite{fulgenzi:inria-00526601} of path planning with social considerations. Only two experiments are presented, both with F-formations in a vis-a-vis formation.

 \subsubsection{Dynamic environments}

 \cite{kollmitz2015time} creates new costmap layers, each representing a predicted future human position, in sequential timesteps. The paper focuses purely on obstacle avoidance, and is thus not as useful when attempting to achieve an interaction pose.

 The SFM \cite{helbing1995social} represents a social space as a set of repulsive forces from humans and obstacles, alongside an attractive force towards a given goal. While the SFM by itself would allow a robot to navigate in a dynamic environment and avoid people, it would not be able to execute an approach without significantly tweaking the force values of any approach target, greatly risking their safety.

The Hybrid Reciprocal Velocity Object (HRVO) \cite{snape2011hybrid}, is the result of successive expansions of the Velocity Object (VO), which by itself represents the set of velocities that, if taken by an agent A, would lead to a collision with an agent B. It is expanded into the Reciprocal Velocity Object (RVO) \cite{van2008reciprocal} which takes into account the reaction of agent B. The HRVO, which takes aspects of the VO and the RVO is purely oriented towards obstacle avoidance.

\cite{truong2017toward} proposed a proactive social motion model that merged the concepts of SFM and HRVO. The merger expands both the SFM and HRVO to take into account not only individuals or objects, but also groups of people, human-object interactions, and human characteristics. The SFM is expanded to generate a repulsive force from these new elements. The HRVO on the other hand generates HRVOs from these new elements, forcing the robot to choose velocities that will not result in disturbing any of them. The two algorithms are then merged by utilizing the velocity that is generated as output from the HRVO as the desired velocity to generate the SFM's attractive force. This algorithm is utilized purely for avoidance and not adapted towards human approach.

\cite{truong2016dynamic} also tackles dynamic environments with the DSZ. An individual's frontal personal space is expanded proportionally to their velocity. \cite{truong2017approach} expands this change to the group space function and alters the approach pose prediction to be capable of taking movement into account. The limit in this work lies in its use of fixed parameters to model personal space.

\section{Approach-oriented adaptive space}
\label{sec: Methodology}

The approach taken in this work expands upon \cite{franciscomelo} taking additional considerations for the safety of humans and adding adaptations geared towards close interaction.

\subsection{Space modeling}

The base personal space model utilized in this work is a modification of the model utilized in \cite{franciscomelo}. While the base 2D Asymmetric Gaussian function proposed in \cite{kirby2010social} considers different standard deviations for the positive and negative x-axis (if centered at the origin), to represent the front and rear of a person, it is still limited to equal deviations on the y-axis. This work alters the function to allow for four different deviations, creating a more flexible model :

$(x_0,y_0)$ is the center of the function;

$A$ is the amplitude of the function;

$\theta_0$ is the orientation of the function;

$\sigma_f$ is the frontal standard deviation ($\theta_0$);

$\sigma_r$ is the rear standard deviation (-$\theta_0$);

$\sigma_{sl}$ is the left standard deviation ($\theta_0 + \frac{\pi}{2}$);

$\sigma_{sr}$ is the right standard deviation ($\theta_0 - \frac{\pi}{2}$).

Algorithm \ref{alg1} computes the value of the Altered Asymmetric Gaussian function, given the parameters, for a cell ($x,y$) given a person or group centered at ($x_0,y_0$).
\begin{algorithm}[!htb]
\caption{Algorithm to compute the altered Asymmetric Gaussian function at a given cell (x,y)}\label{alg:alg1}
\begin{algorithmic}[1]
\STATE \textbf{function} {\textsc{AlteredAsymmetricGaussian}}($x,y,x_0,$\\$y_0,\theta_0,A,\sigma_f,\sigma_r,\sigma_{sl},\sigma_{sr}$):
\STATE \hspace{0.5cm}$\theta \longleftarrow atan2(y-y_0,x-x_0)$;
\STATE \hspace{0.5cm}$\alpha \longleftarrow \theta - \theta_0$;
\IF{$abs(\alpha) < \frac{\pi}{2}$}
    \STATE $\sigma_x \longleftarrow \sigma_f$
\ELSE
    \STATE $\sigma_x \longleftarrow \sigma_r$
\ENDIF
\IF{$\alpha < 0$}
    \STATE $\sigma_y \longleftarrow \sigma_{sr}$
\ELSE
    \STATE$\sigma_y \longleftarrow \sigma_{sl}$
\ENDIF
\STATE $d \longleftarrow \sqrt{(x-x_0)^2+(y-y_0)^2}$
\STATE\textbf{return} $A \exp\left(-\left(\left(\frac{d \cos(\theta-\theta_0)}{2\sigma_x}\right)^2+\left(\frac{d \sin(\theta-\theta_0)}{2\sigma_y}\right)^2\right)\right)$
\end{algorithmic}
\label{alg1}
\end{algorithm}

Given a person $p_i$ positioned at ($x^P_i,y^P_i$) and oriented towards $\theta^P_i$ it is possible to calculate the value of their personal space for a position ($x,y$) through the use of Algorithm \ref{alg1}, if supplied the parameter set [$A^P,\sigma^P_{i\_f},\sigma^P_{i\_r},\sigma^P_{i\_sl},\sigma^P_{i\_sr}$], specific to the person in question. These parameters will affect the shape and size of the personal space.

With a set of N individuals, the function that represents the merged personal space of every individual can thus be represented by:
\begin{equation}
    F^P(x,y) = max(f^P_1(x,y),...,f^P_N(x,y)),
\end{equation}
with $(f^P_1(x,y),...,f^P_N(x,y))$ representing the personal space functions of each individual.

Groups of humans, can be divided into the spaces described in Section \ref{Proxemics}. This work also considers another space, introduced in \cite{franciscomelo} named the group space, limited by a radius defined as the average of the distances of every group member from the group center. Group space is modelled utilizing the Altered Asymmetric Gaussian. A group $k$ itself can be defined through the parameters $g_k = (x^g_k,y^g_k,r^g_k, \theta^g_k)$ where $(x^g_k,y^g_k)$ represents the center of the group, $r^g_k$ is the group radius, and $\theta^g_k$ is the orientation of the group, calculated as the average of the orientations of the velocity vector of every group member. This is modeled through  the mentioned Gaussian with its center at the center of the group and defined by the parameters $[A^g, \sigma^g_{k\_f}, \sigma^g_{k\_r}, \sigma^g_{k\_s}]$. $\sigma^g_{k\_s}$ represents both side deviations as they do not differ in this case.

With a set of N groups, the function that represents the merged group space of every group is computed through:
\begin{equation}
    F^g(x,y) = max(f^g_1(x,y),...,f^g_N(x,y)),
\end{equation}
with $(f^g_1(x,y),...,f^g_N(x,y))$ representing the group space of every group.

By joining the function that computes every personal space and this function, the model of every individual and group within the space is obtained:
\begin{equation}
    F(x,y) = max(F^P(x,y),F^g(x,y))
\end{equation}
\subsection{Adaptive space}
The space model adaptations developed in this dissertation can be divided into three different parts: (i) individual adaptation, (ii) group adaptation, and (iii) velocity adaptation.

\subsubsection{Individual adaptation}

The adaptation applied to an individual is based on the idea that a robot must approach a human being closer than the edge of its personal space in order to properly interact with them. As most approaches to lone humans should be done from within their field of view (and thus from their front) \cite{walters2007robotic,woods2006methodological} the only adaptation applied to their personal space model is to lower the value of the frontal standard deviation $\sigma^P_f$ by a preset value $\zeta$. The change is thus simply computed through this calculation, for a given approach target $i$:
\begin{equation}
    \sigma_{aux} = max(0.45,\sigma^P_f-\zeta),
\end{equation}
\begin{equation}
    \sigma^P_{i\_f} = min(\sigma^P_f,\sigma_{aux}),
\end{equation}
where the lowered value is compared to the threshold of the intimate space as defined by Hall \cite{ehall1990} in order to guarantee the sanctity of the person's intimate space.

\subsubsection{Group adaptation}

The adaptation to group space is done not to the parameters of the group model itself, but to the parameters of each group member individually. The adaptation starts from the parameters provided by the previous algorithm \cite{franciscomelo} and attempts to determine if it is possible to lower the values of the lateral standard deviations of each personal space model. Each personal space calculation requires, as input, the Gaussian parameters of an individual $\sigma^P = $($A^P,\sigma^P_{f},\sigma^P_{r},\sigma^P_{sl},\sigma^P_{sr}$), their pose ($x^P,y^P,\theta^P$) and only these last parameters from the individuals to their left ($x^P_l, y^P_l,\theta^P_l$) and to their right ($x^P_r, y^P_r,\theta^P_r$) in the group. In a group of two individuals only one of these will be considered.

In order to prevent collisions between the robot and any of the individuals, points are calculated to the side of each individual, in a line perpendicular to their orientation, at a distance ($s_h$) determined by user inputted values.

The auxiliary points are calculated as such:
\begin{equation}
    A_{left} = \Bigl(x^P+s_h\cdot\cos\left(\theta^P+\frac{\pi}{2}\right),y^P+s_h\cdot\sin\left(\theta^P+\frac{\pi}{2}\right)\Bigr),
\end{equation}
\begin{equation}
    A_{right} = \Bigl(x^P+s_h\cdot\cos\left(\theta^P-\frac{\pi}{2}\right),y^P+s_h\cdot\sin\left(\theta^P-\frac{\pi}{2}\right)\Bigr),
\end{equation}
\begin{equation}
    A_{l\_adj} = \Bigl(x^P_l+s_h\cdot\cos\left(\theta^P_l-\frac{\pi}{2}\right),y^P_l+s_h\cdot\sin\left(\theta^P_l-\frac{\pi}{2}\right)\Bigr),
\end{equation}
\begin{equation}
    A_{r\_adj} = \Bigl(x^P_r+s_h\cdot\cos\left(\theta^P_r+\frac{\pi}{2}\right),y^P_r+s_h\cdot\sin\left(\theta^P_r+\frac{\pi}{2}\right)\Bigr).
\end{equation}
The algorithm then calculates the distance between each pair of points and determines whether there is space for the robot to insert itself in between, by comparing the distances to a provided value for the lateral dimension of the robot ($s_r$). The distances calculated are euclidean:
\begin{equation}
    d_{left} = \sqrt{(A_{left} - A_{l\_adj})^2}
\end{equation}
\begin{equation}
    d_{right} = \sqrt{(A_{right} - A_{r\_adj})^2}
\end{equation}
It also calculates whether the difference in orientation between two individuals is set between two different user defined values. Should the answer be positive, it will determine whether the current lateral standard deviations permit an approach and adapt them should they not. This is done by calculating a point ($B_{aux}$) at the minimum distance from the relevant auxiliary point ($A_{left}$ or $A_{right}$), that complies with the space requirements, projecting it on a line perpendicular to the person's direction, and then measuring the distance between the projection and the person. Should this distance be less than the preexisting standard deviation, then it is replaced. This algorithm can be described as such, for the case between an individual and the one on their left:
\begin{equation}
    \mathbf{v_{aux}} = \frac{(A_{left} - A_{l\_adj})}{d_{left}},
\end{equation}
\begin{equation}
    B_{aux} = A_{left} + \frac{(d_{left}-s_r)}{2}\cdot \mathbf{v_{aux}},
\end{equation}
\begin{equation}
    d_{aux} = \sqrt{((x^P,y^P) - P_{proj})^2},
\end{equation}
\begin{equation}
    \sigma^P_{sl} = min(\sigma^P_{sl},d_{aux}),
\end{equation}
\begin{equation}
    \sigma^P_{sl} = max(\sigma^P_{sl},0.225),
\end{equation}
where $P_{proj}$ is the projection of $B_{aux}$ on the line perpendicular to the person's orientation. The same process is done for the right side to obtain $\sigma^P_{sr}$. $0.225$ represents the average value for half the lateral size of a human.

\subsubsection{Velocity adaptation}
\label{Velocity space}

It is necessary to adapt the space models to take into account human velocity. The choice made in this work is to increase the frontal standard deviation. The adaptation has three parameters: the adaptation factor ($a_{adapt}$), the maximum adaptation limit ($a_{limit}$) and the distance limit ($d_{limit}$). The first parameter determines the proportion by which the velocity affects the standard deviation. The second one provides an upper limit for the adaptation. The distance limit lowers the adaptation as the robot approaches the individual or group. The same algorithm is applied to groups and individuals and thus it will only be described once with the input parameters ($x^{Pg},y^{Pg},\sigma^{Pg}_f$) to represent the position and the frontal standard deviation. A group's velocity is calculated by averaging the velocity of all its members.
\begin{equation}
    d_{mod} = 
    \begin{cases}
    min(1,\frac{dist}{d_{limit}}\cdot2) & \text{if } d <= d_{limit} \\
    1 & \text{if } d > d_{limit}
    \end{cases}
\end{equation}
\begin{equation}
    \sigma^{Pg}_f = min\Bigl(\sigma^{Pg}_f+a_{limit}, \sigma^{Pg}_f\cdot(1+  d_{mod}\cdot a_{adapt}\cdot v_{mag})\Bigr)
\end{equation}
where $d$ is the distance between the robot and an individual or a group and $v_{mag}$ is the magnitude of the human velocity.
\subsection{Approach pose estimation}
\label{ApproachPose}

Utilizing the space models calculated through the algorithms presented in the previous section, it is possible to estimate possible approach poses. The algorithm utilized is mostly based on the one defined in \cite{franciscomelo} with changes and additions that will be described in this section. These changes can be separated into two types: (i) safety and comfort oriented changes, and (ii) dynamic environment changes.

\subsubsection{Safety and comfort}

The changes in this section intend to enable a safe and comfortable approach for the people being approached. The first change is to attempt an approach within the field of view of all involved people. This is done by filtering the approach circumference for areas within the common field of view of every individual. Should no valid zones be detected with this condition, the algorithm defaults to the original filtering behaviour, ignoring the field of view as a factor.

The preexisting algorithm does not check if the robot fits in the approach zone. With a preset robot size as input alongside a list of all approaching zones found, the width of the zones themselves is evaluated by calculating the distance between the two farthest points within the zone, and that distance is then compared to the preset size. Should the distance be greater, then this zone should be considered valid for approach. The approach pose chosen is the one closest to the robot itself, of all the valid ones.

\subsubsection{Dynamic environment adaptation}

The first change in this section mirrors Section \ref{Velocity space}. The maximum limit the approach circumference is allowed to expand to is raised proportionally to the magnitude of the individual's or group's velocity, limited by a preset value. This change is also more limited the closer the robot is to the approach target. All equations applied in Section \ref{Velocity space} apply to this section with the standard deviation replaced by the maximum limit of the approach circumference radius. For the purposes of differentiation, the approach pose equivalents of the variables presented in \ref{Velocity space} are named: $d_{a\_limit}$, $a_{a\_limit}$, and $a_{a\_adapt}$.

A dynamic environment grants the robot less time for calculations, as the movement of a human quickly renders them obsolete. The robot needs to quickly evaluate possible approach positions. This is done by increasing the expansion step taken when no valid approach zone is found proportionally to the velocity. This can be defined as such:
\begin{equation}
    step = max(step, step\cdot v_{mag}\cdot v_{mod}),
\end{equation}
where the $v_{mod}$ is a user defined value that determines how much the velocity affects the step.

The last change applies to the field of view defined in the previous subsection. The expanded space model leads to farther and thus wider approach zones. This could lead to lateral approach poses being chosen, which could be suboptimal. It is thus necessary to narrow the field of view the farther away the current approach circumference is from its initial radius. This can be calculated as such:
\begin{equation}
    fov = \frac{f_{ifov}}{f_{mod}\cdot\frac{approach\_radius}{group\_radius}},
\end{equation}
where $f_{ifov}$ is the initial angle of the field of view and $f_{mod}$ is a user defined value that determines how much the distance from the initial radius narrows the field of view.

\subsection{Results}
\label{Resultschap3}

In this section, a preliminary study of the space model adaptation algorithm is done, through a comparison to the results obtained when utilizing the algorithm present in \cite{franciscomelo}. The objective is to evaluate, on an initial basis, how the algorithm performs when utilized on existing datasets and its performance when compared with preexisting work.

\subsubsection{Visualization}

\begin{figure}[!htb]
    \centering  
    \includegraphics[width=0.45\textwidth]{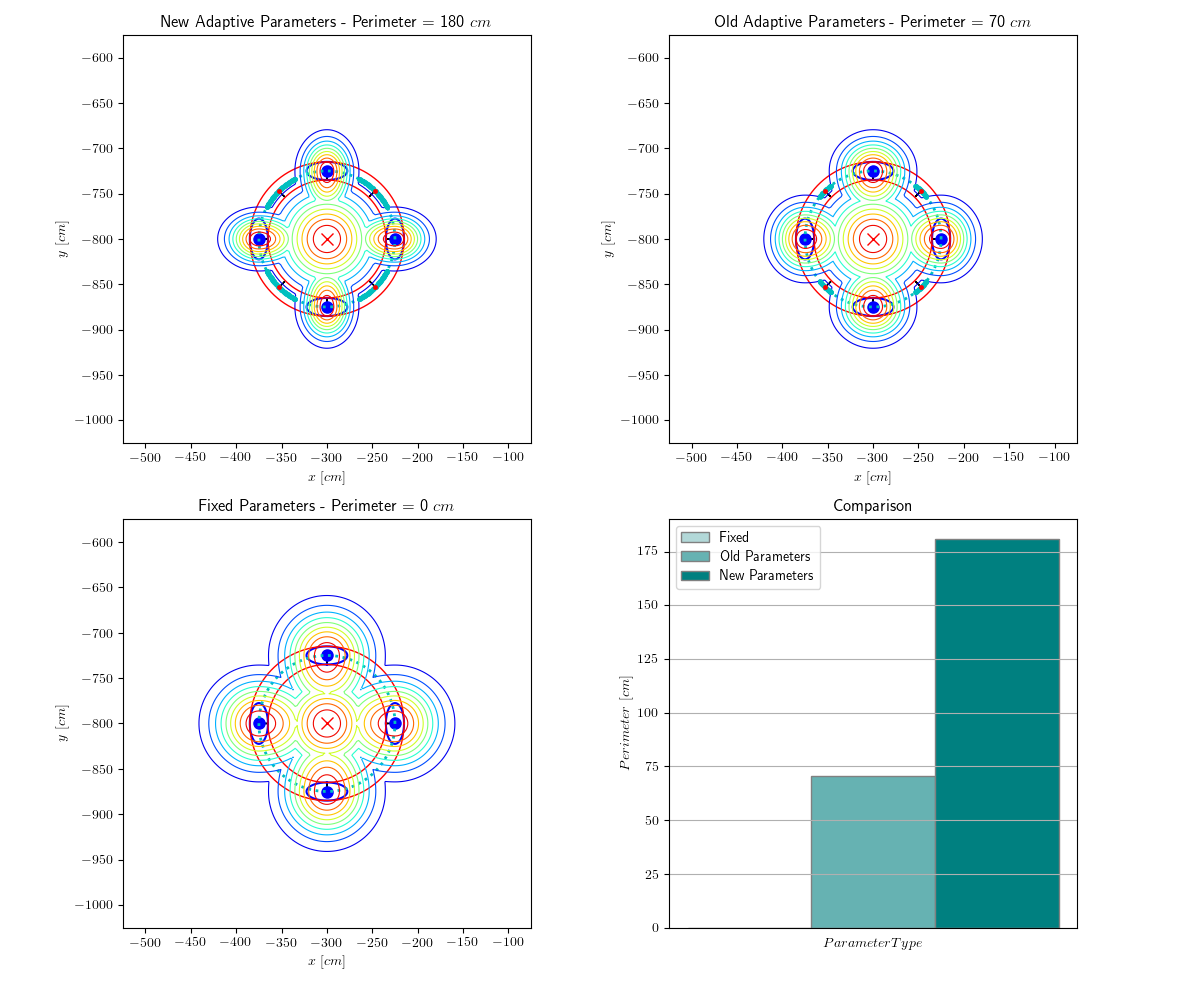}
    \caption{Visualization of the group space and personal space pertaining to Group 17 of the Groups Data dataset, alongside possible approach poses. This group's members are all at a distance of $0.75m$ from the center of the group.}  
    \label{Group17}
\end{figure}

A dataset, comprised of 17 situations, was utilized for an initial evaluation. Due to the nature of the algorithm, not all configurations will result in an adaptation, with only  8 of the 17 being adapted. Successful situations tend to be those with enough members that a space model adaptation is justified, but also enough free space in between each individual to guarantee a safe approach. Three of the cases present in the dataset are variations of the same situation: 4 people placed equidistantly from each other and at the same radius from the center of the group with varying distances from the center. These distances are $0.5m$, $0.75m$ and $1m$. Only the second case was successfully adapted. The first case failed due to the individuals being too close together, creating a situation in which there were no safe approach poses. In the third case, the individuals were separated enough that safe approach poses existed by default and nothing was adapted. In the second case, depicted in Fig. \ref{Group17}, the individuals were far enough from each other that safe approach poses could be determined, but the original space model blocked these, leading to the adaptation by the new algorithm.

\subsubsection{Comparative Study}

For a more formal comparative study of the adaptation algorithm, it is necessary to utilize larger datasets to obtain representative results. The two datasets utilized were the same as those used in the comparative work: \textbf{Synthetic Data} \cite{cristani2011social} and \textbf{IDIAP Poster Data} \cite{hung2011detecting}. Table \ref{datasets} describes the composition of the groups present within the datasets.
\begin{table}[!htb]
\caption{Composition of the datasets}
\label{datasets}
\centering
\begin{tabular}{|l|llllll|l|}
\hline
\multirow{2}{*}{Dataset} & \multicolumn{6}{l|}{Group Size}                                                                                                      & \multirow{2}{*}{Total} \\ \cline{2-7}
                         & \multicolumn{1}{l|}{2}   & \multicolumn{1}{l|}{3}   & \multicolumn{1}{l|}{4}  & \multicolumn{1}{l|}{5}  & \multicolumn{1}{l|}{6} & 7 &                        \\ \hline
Synthetic                & \multicolumn{1}{l|}{180} & \multicolumn{1}{l|}{80}  & \multicolumn{1}{l|}{-}  & \multicolumn{1}{l|}{-}  & \multicolumn{1}{l|}{-} & - & 260                    \\ \hline
IDIAP Poster             & \multicolumn{1}{l|}{152} & \multicolumn{1}{l|}{106} & \multicolumn{1}{l|}{56} & \multicolumn{1}{l|}{20} & \multicolumn{1}{l|}{5} & 6 & 345                    \\ \hline
\end{tabular}
\end{table}

Both algorithms were fed the datasets as input and an analysis of the resultant approach perimeter was done. Several iterations of the algorithm were executed in order to allow for a test of various possible scenarios and tolerances, resulting from the variability of the algorithm's input parameters. These iterations are defined by changes in the $s_r$ and $s_h$ parameters representing the size of the robot that would be expected to approach the groups and the desired minimum safety tolerance for the distance between the robot and the person, respectively. $s_r$ has two values, $0.45m$ and $0.8m$. These values represent, respectively, a roughly human-sized robot, and a size approaching that of the Vizzy robot. $s_h$ varies from $0.225m$ to $0.45m$, the minimum allowed value for the space model per the original algorithm and the value for intimate space radius per Hall, respectively.

\begin{figure}[!htb]
    \centering
    \begin{minipage}[b]{0.2\textwidth}
        \includegraphics[width=\textwidth]{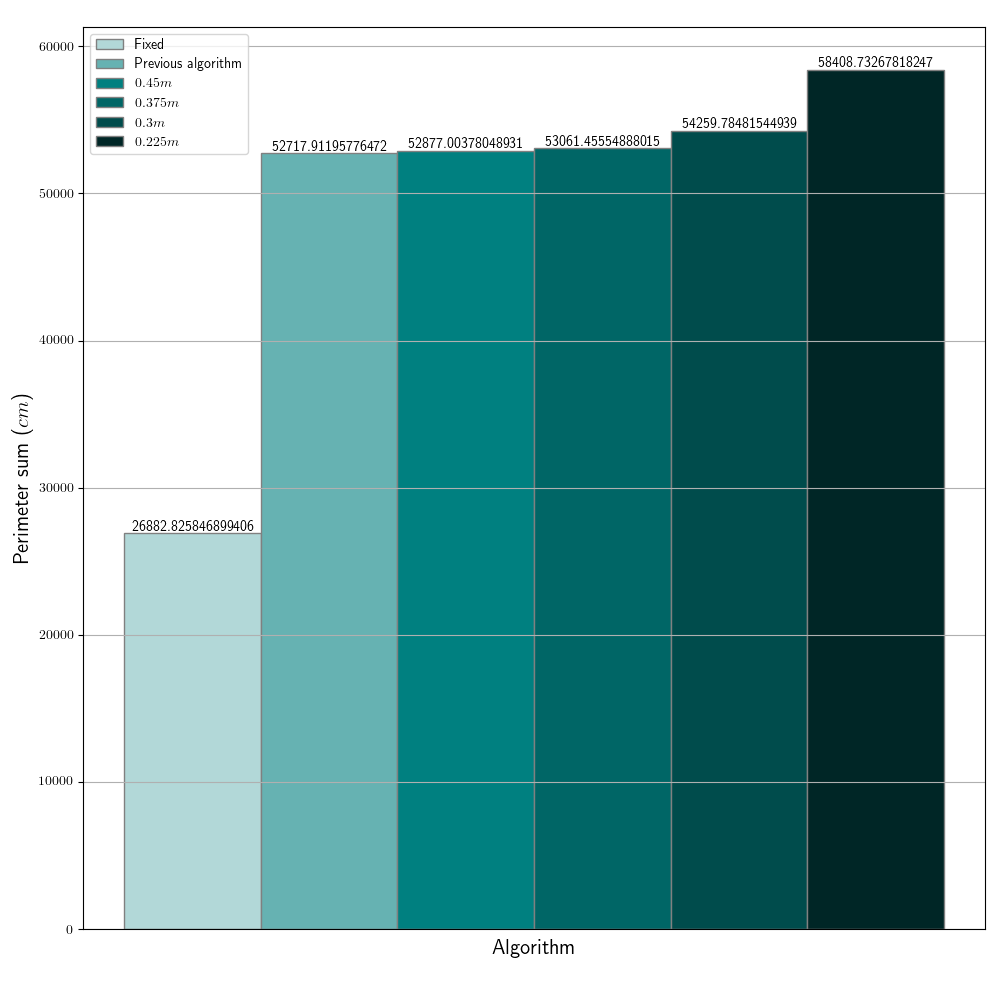}
        \caption{Total perimeter sum for Human-sized robot - IDIAP dataset}
        \label{IDIAPHumanTotal}
    \end{minipage}
    \hfill
    \begin{minipage}[b]{0.2\textwidth}
        \includegraphics[width=\textwidth]{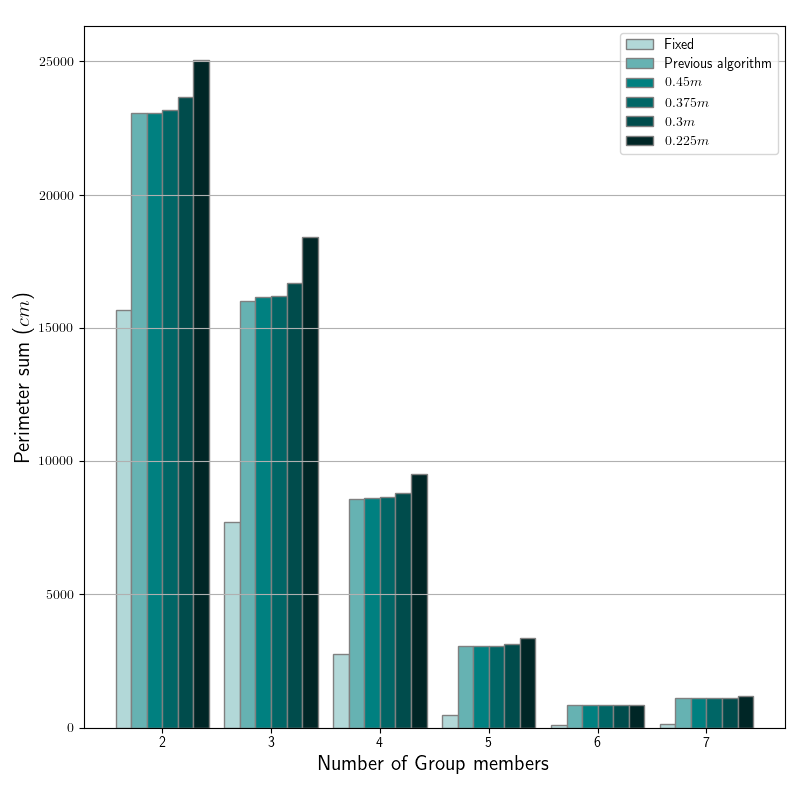}
        \caption{Perimeter sum divided by number of group members for Human-sized robot - IDIAP dataset}
        \label{IDIAPHumanGroup}
    \end{minipage}   
\end{figure}

\begin{figure}[!htb]
    \centering  
    \includegraphics[width=0.4\textwidth]{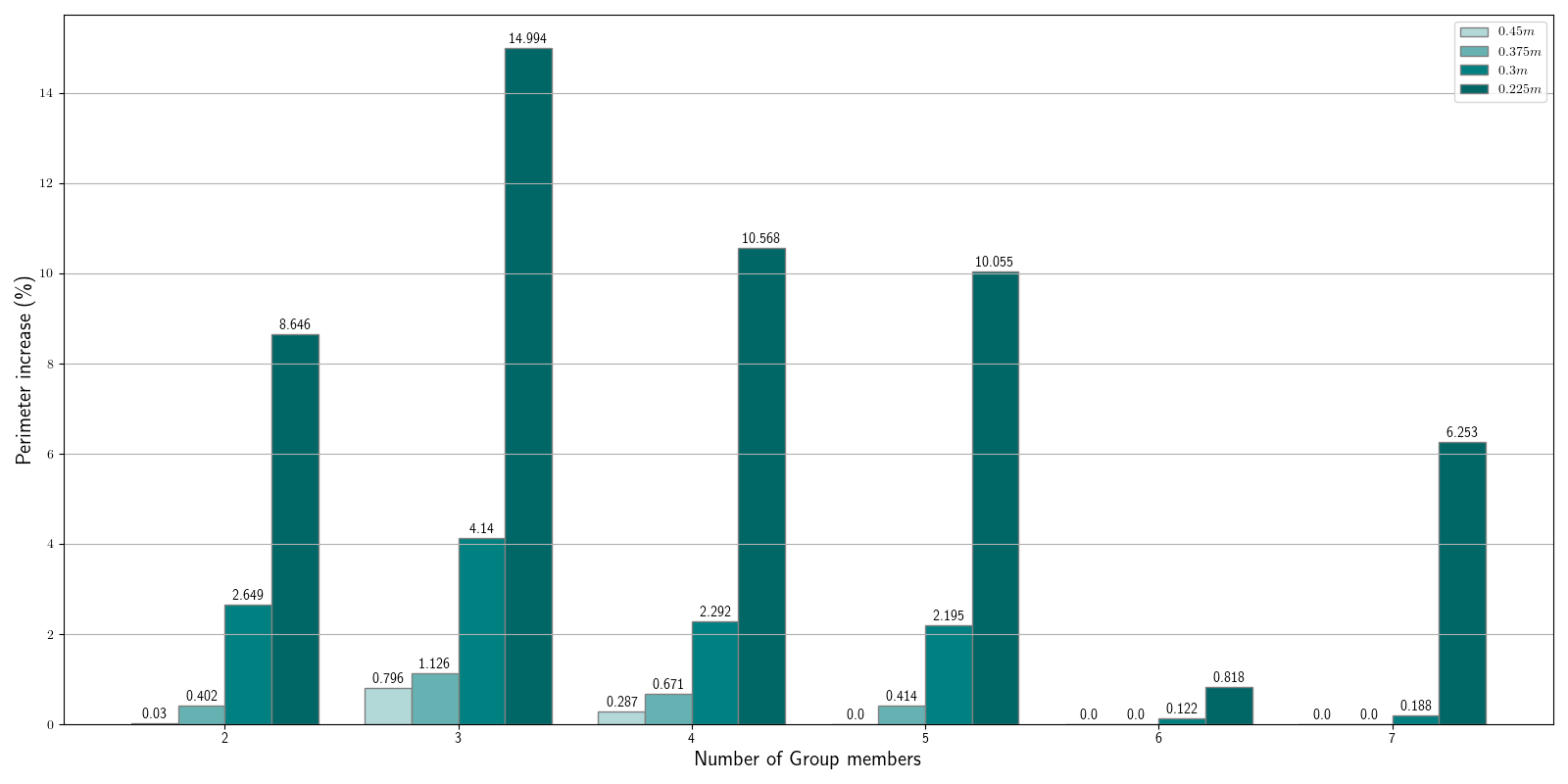}
    \caption{Percentage of increase in perimeter sum compared to old algorithm in Human-sized robot - IDIAP dataset}   
    \label{IDIAPHumanPercentage}
\end{figure}

Figs. \ref{IDIAPHumanTotal} to \ref{IDIAPHumanPercentage} represent the results of the experiment for a human-sized robot with the \textbf{IDIAP Poster Data} dataset. As can be observed by analysing the plots, the algorithm developed in this work does increase the approach perimeter in comparison to the original algorithm. As would be expected, loosening the tolerance in the distance allowed between the robot and the human results in a larger perimeter. At its greatest increase with $s_h = 0.225m$, the algorithm has a total increase in perimeter of $10.7\%$. As the situation with the starkest differences, it is this minimum tolerance experiment that will be analysed in depth, though all the rest follow its trends and thus similar results can be extrapolated.

By observing Fig. \ref{IDIAPHumanPercentage}, it is possible to note that the increases were not uniform for every group type. Groups with three members were the ones most affected at $15\%$, followed by groups of four, five, two, seven and lastly six members. The groups with six and seven members can not be utilized to provide any viable conclusions, due to the low number of samples. It is apparent that groups of only two members had the smallest percentage increase of these 4 sizes. This is due to groups of two members being unlikely to be positioned in a manner in which adaptation is viable or necessary. Most of the common shapes for these groups (vis-a-vis or side-by-side and similar) would not result in any adaptation. The remaining common forms of group arrangement (L-shape, C-shape or V-shape) often will also not result in adaptation as the available approach pose is already large enough.

Groups of three members can be concluded to be the most viable for adaptation. This could be seen as a combination of having enough members to avoid the situation verified with groups of two members, tending more towards circular formations, alongside having enough free space, due to having less members than groups of four or five people, for the algorithm to find viable spaces to adapt.

The other three scenarios (\textbf{IDIAP Poster Data} dataset with a Vizzy-sized robot and \textbf{Synthetic Data} dataset with Human and Vizzy-sized robots) present results that possess similar trends to the ones already presented with justifiable variations. The \textbf{IDIAP Poster Data} dataset with a Vizzy-sized robot results in a smaller total increase in perimeter with a $5.4\%$ increase, due to the larger robot being able to safely approach a smaller amount of groups. The cases utilizing the \textbf{Synthetic Data} dataset have even smaller increases at $5\%$ and $3.6\%$, for the human and Vizzy sized robots respectively, due to the larger number of groups with two members and lesser number of groups with 3 members, which following the previous conclusions is a justifiable result.

\section{Socially Reactive Navigation System}
\label{sec:Experiments}

It is necessary to implement the space model and approach pose estimation and integrate them with the robot's navigation system. A socially aware navigation system runs parallel to the classic robot navigation framework. The usual robot navigation framework is composed of four blocks: perception, localization, motion planning and motion control. The socially aware navigation system adds the following four blocks: Human detection and tracking, group detection, modeling of space and approach pose estimation. These last two blocks correspond to the main contribution of this work. The system was implemented utilizing the Robot Operating System (ROS).

\subsection{Detection of Social Scenarios}

\subsubsection{Human Detection}

It is necessary for the robot to be able to detect humans in order to properly integrate them into the navigation system. This work makes use of OpenPose \cite{cao2021openpose}, a real time multi-person 2D pose detection, system to identify humans. It detects humans by identifying body parts and associating them with individuals via a non-parametric representation named Part Affinity Fields. Utilizing the 2D pose, a method developed in \cite{joaoavelino} is then capable of utilizing homography to determine the 3D pose.

\subsubsection{Group Detection}

To properly address groups of people it is then necessary to be able to identify them, given the poses of people in the environment. \cite{kollakidou2021enabling} proposes a hierarchical clustering method to identify groups, taking as input the people's poses and outputting the identified groups.

\subsection{Space Modeling}

This work makes use of the $costmap\_2d$ ROS package to implement the space models described previously. It utilizes the layered costmap method and considers the following standard layers: (i) \textit{Static Layer}, (ii) \textit{Obstacles Layer}, and (iii) \textit{Inflation Layer}. This work also makes use of two custom layers developed in \cite{franciscomelo}, the \textit{Clean People Layer} and the \textit{Adaptive Layer}. The first of these removes people from the costmap before inflation by the \textit{Inflation Layer}, while the \textit{Adaptive Layer} implements the space models themselves. Contrary to the original implementation of this last layer, the human body isn't marked entirely as lethal, to guarantee the proper decay of the Gaussian function that represents the space model.

The layers are implemented in the same order as in \cite{franciscomelo}: (i) \textit{Master}, (ii) \textit{Adaptive Layer}, (iii) \textit{Inflation Layer}, (iv) \textit{Clean People Layer}, (v) \textit{Obstacles Layer}, and (vi) \textit{Static Layer}.

\subsection{Approach Pose Estimation}

The approach pose algorithm is implemented as described in \ref{ApproachPose}. The approach poses are checked by verifying the cells in the group radius. A cell is marked as free to approach should the value be below a certain threshold defined by the user, being added to the possible approaching area. A tracking of the desired approach target is also done, for lack of software that can keep constant track of specific individuals or groups, by checking the closest group or individual to the original target and checking that they're located under a threshold distance.

\subsection{Evaluation}

In order to properly evaluate some of the presented scenarios it is necessary to utilize criteria that is geared towards analysing social scenarios. To this end a set of Human Safety and Comfort Indexes (HSCIs)\cite{truong2017approach} were chosen. These indexes evaluate human comfort and approach direction. These indexes are: (i) Social Individual Index (SII), (ii) Social Group Index (SGI), and (iii) Social Direction Index (SDI).

The higher the values of SII and SGI then the closer a robot is to breaching the personal or group spaces, respectively. Nonetheless, certain higher values could be desired as it shows a robot is capable of approaching humans enough for closer interaction without causing discomfort. Greater values of SDI simply signify that the robot's and the human's orientation is closer to being aligned. \cite{truong2017approach} defined a value of $0.14$ as the upper threshold that the SII and SGI should not surpass to guarantee the comfort of a human.

\subsection{Results}

This section seeks to evaluate the performance of the algorithms through several scenarios, both simulated and through a real robotic platform to demonstrate that the robot is capable of navigating a social environment and safely interact with humans. While the static scenarios will be evaluated through use of the real robot, due to hardware and software constraints the dynamic scenarios must be evaluated through simulation.

Three ROS packages, originally developed in \cite{franciscomelo} were altered to implement the algorithms as presented in this work.

\subsubsection{Simulation Experiments}

The simulations were run in the Gazebo simulator, utilizing the robotic platform Vizzy \cite{moreno2016vizzy}. The simulation environment is an empty replica of the 7th floor of the North Tower at ISR Lisbon. The parameters utilized to initialize the original adaptive space algorithm were equal to those utilized in Melo's \cite{franciscomelo} own simulation experiments, with the exception of the amplitude of the Gaussian function which was set to 211. The following parameters were defined for the experiments: $\zeta = 0.3m$, $s_h = 0.375m$, $s_r = 0.8m$, $d_{limit}=d_{a\_limit}=6m$, $a_{adapt} = a_{a\_adapt} = 1.5$, $a_{limit} = 1m$, $a_{a\_limit} = 1.2m$, $v_{mod} = 10$, $f_{ifov} =90^\circ$, and $f_{mod} = 1.1$. The simulations were run on a computer with a 2,8 GHz Quad-Core Intel Core i7 processor and 16 GB 2400 MHz DDR4 of RAM running Ubuntu 18.04 and using ROS Melodic. The simulator was Gazebo 9.

\textbf{Static Environments}: The models representing people remain in place throughout the entire duration of these experiments, with the robot being the only mobile element as it executed an approach. The experiments were divided into an individual approach experiment and a group approach experiment, and each would be run twice, first without space model adaptation and the second time with. 

The individual experiment consisted simply of a single human model placed in front and to the right of the robot, near one of the room's walls, which the robot was then commanded to approach.

\begin{figure}[!htb]
    \centering
    \includegraphics[width=0.45\textwidth]{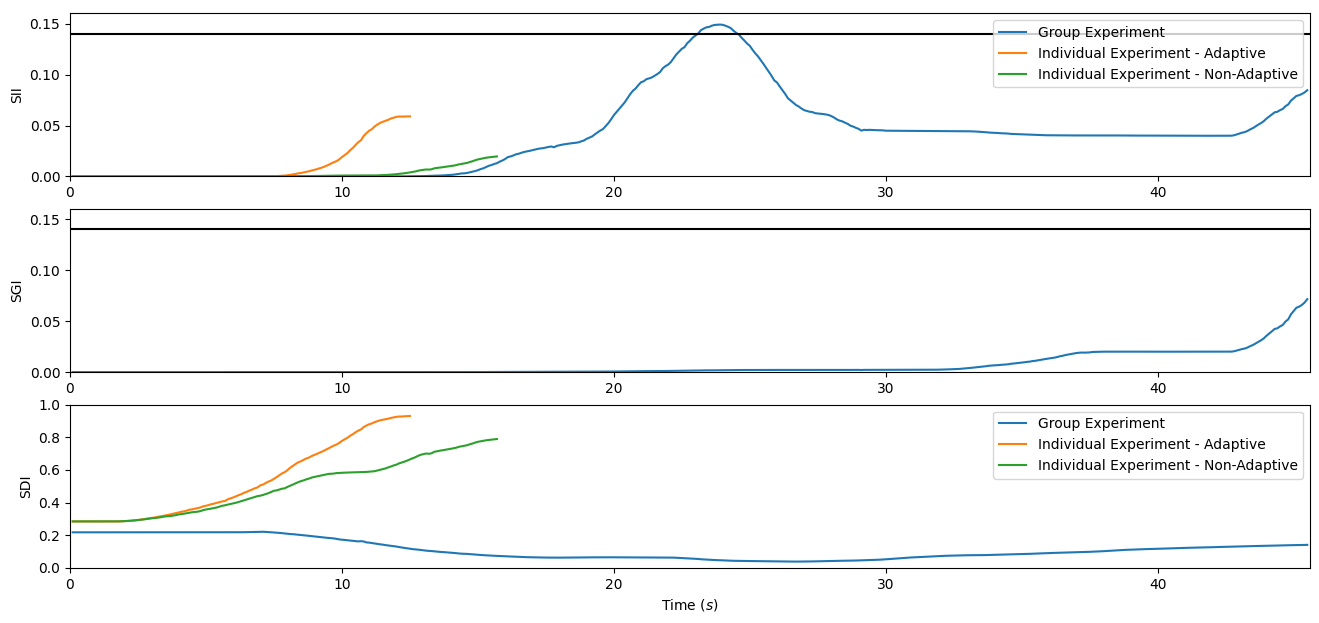}
    \caption{Human Safety and Comfort Indexes from the static simulation experiments}
    \label{StaticIndexes}
\end{figure}

The only significant difference in the results is the closer approach of the adapted experiment, enabling more direct and involved interaction. These results are corroborated by the HSCIs extracted during the experiments in Fig. \ref{StaticIndexes} where it is possible to see that the SDIs are similar, but the final value of the SII in the iteration with space model adaptation is higher than the one without, but still below the comfort threshold.

In the group experiment, two groups of 4 people were created and the robot was commanded to approach one of them. In the case without adaptation, no approach pose was identified as valid. Utilizing this work's adaptation, the space model was changed enough to allow for a safe approach. From an analysis of the HSCIs in Fig. \ref{StaticIndexes} of the successful approach it is possible to see that the indexes were kept below the comfort threshold at all times apart from a moment in the middle of the approach. The SII went slightly over the comfort threshold when the robot moved behind one of the group members on the path to the approach pose. This breach is less worrying than a usual one, as humans are less likely to notice breaches of their personal space when these happen behind them.

\textbf{Dynamic Environments}: Simulations in a dynamic environment were divided into two different scenarios: approaching a lone person, or a group of two people side-by-side. In both of these scenarios, the human models move forward at a constant velocity until they reach a specific distance threshold to the robot. All elements of these experiments are controlled via either an automated script or the ROS navigation. Two different scenarios are contemplated both for the individual and group experiments: at the lowest velocity the robot begins its approach at an angle to the humans, while in the two highest velocities the robot begins directly in front of them.

In order to properly compare the performance of the velocity adaptation algorithms against the algorithm without these, each scenario was repeated four times with a different configuration of the algorithms used. The configurations are as follows: (i) space model and approach pose estimation without velocity adaptation, (ii) space model without adaptation and approach pose estimation with adaptation, (iii) space model with velocity adaptation and approach pose estimation without adaptation, and (iv) space model and approach pose estimation with velocity adaptation.

The results from the first two configurations tend to be rather similar, as they're effectively running the same algorithm. On the other hand, the space model adaptation without approach pose estimation adaptation is completely nonfunctional as the estimation is incapable of consistently finding a proper approach pose without being adapted too. The configuration with both adaptations has a more consistent performance than the other three. This is noticeable in both individual and group experiments. Videos of these experiments are available.\footnote{Individual Approach Experiment -  \url{https://youtu.be/XZ7G7S1gUDk}} \footnote{Group Approach Experiment - \url{https://youtu.be/gjiD-j1-r44}}

\begin{figure}[!htb]
    \centering  
    \includegraphics[width=0.5\textwidth]{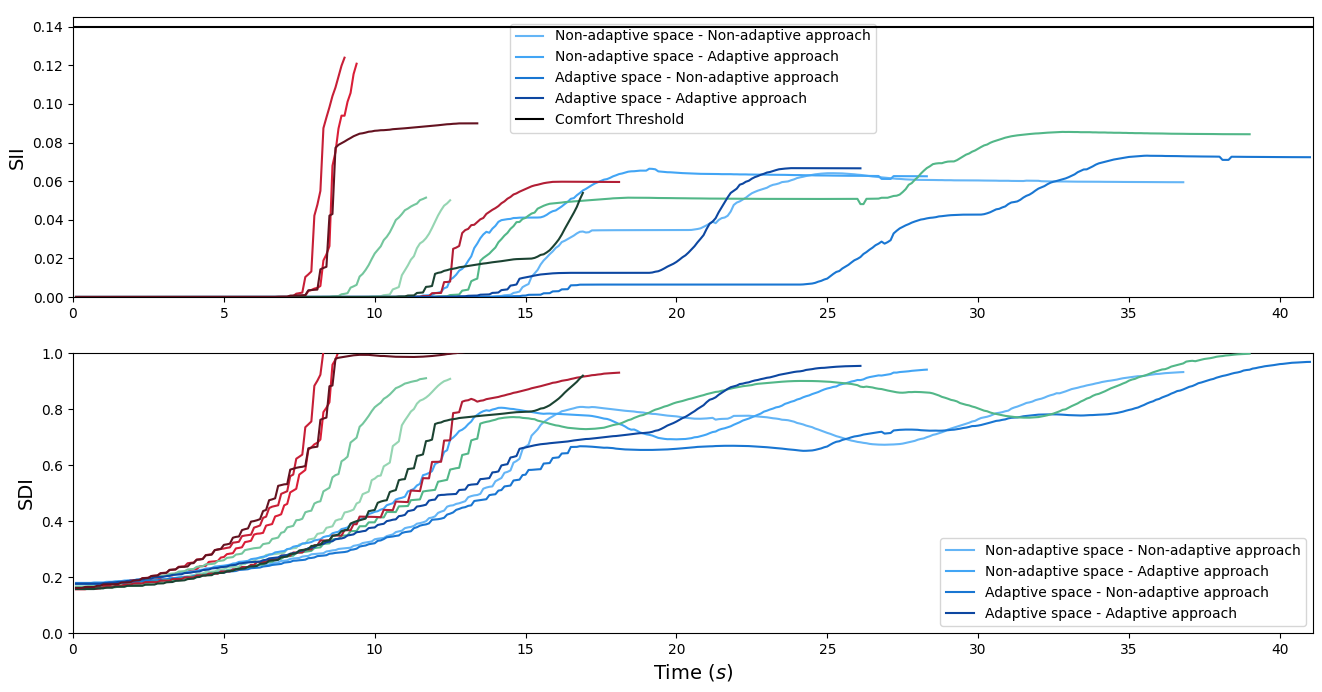}
    \caption{Human Indexes for the dynamic individual approach experiment. Blue represents the lateral approach at $0.5m/s$, while green and red are the frontal approaches, at $1m/s$ and $1.5m/s$ respectively. The colour progression, from lighter to darker is equal for all experiments.}
    \label{DynamicIndiv}
\end{figure}

\begin{figure}[!htb]
    \centering  
    \includegraphics[width=0.5\textwidth]{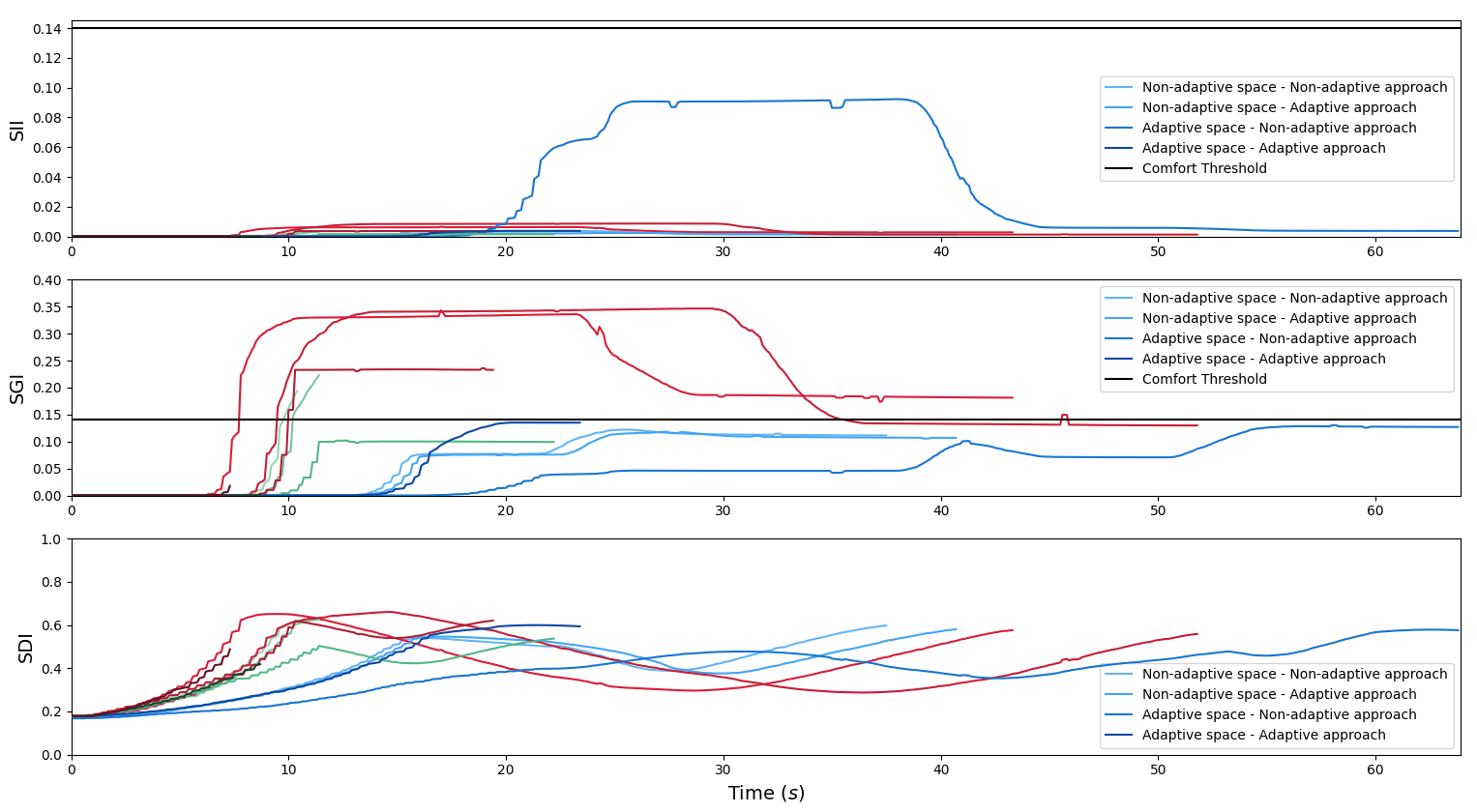}
    \caption{Human Indexes for the dynamic group approach experiment. Blue represents the lateral approach at $0.5m/s$, while green and red are the frontal approaches, at $1m/s$ and $1.5m/s$ respectively. The colour progression, from lighter to darker is equal for all experiments.}
    \label{DynamicGroup}
\end{figure}

The HSCIs presented in Figs. \ref{DynamicIndiv} and \ref{DynamicGroup} corroborate the conclusions from the previous paragraph, proving the greater reliability of the algorithm developed in this work.

\subsubsection{Real life experiments}

Real life experiments are separated into two sections: experiments run with a single individual and experiments run with a group. In order to run the experiments, all the algorithms were implemented on the previously mentioned robotic platform, Vizzy \cite{moreno2016vizzy}.

Individual experiments were divided into two similar parts, though one evaluated preference while the other served to evaluate distance.

The preference part of the experiment made use of 19 volunteers (12 male, 6 female and 1 that preferred not to say). They were asked to stand in front of the robot, and were given a map. They were instructed to wait for the robot to approach them and, once it stopped, to show it the map as though it were seeking instruction on directions. Upon finishing this task the participants were told to fill in a questionnaire consisting of a Godspeed style questionnaire \cite{bartneck2009measurement}, composed of 10 questions on an altered scale of 1 to 9, and then the task was repeated. They would then be faced with the same questionnaire and a final question which asked in which of the two experiments was it easier to accomplish the task they were given, with the option to answer 'Experiment 1', 'Experiment 2' or 'Neither'. The first iteration utilized the non-adaptive space model and approach pose estimation, while the second used the algorithms developed in this work.

\begin{figure}[!htb]
    \centering  
    \includegraphics[width=0.475\textwidth]{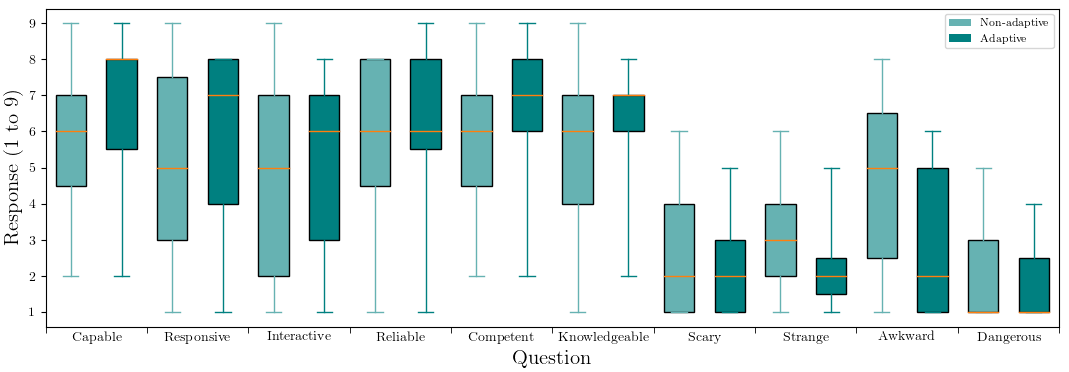}
    \caption{Responses to the Godspeed-style Questionnaire}
    \label{GodspeedAnswer}
\end{figure}

The results of the Godspeed Questionnaire are displayed in Fig. \ref{GodspeedAnswer}. These results were then compiled and statistic tests were applied to decide whether any of the results presented a significant statistic difference. The T Test and Mann-Whitney U test were utilized. The results of these tests are displayed in Table \ref{GodspeedTests}. The null hypothesis being tested is that the distribution of the samples belonging to the two experiments is the same, and thus possesses no relevant statistical difference.

\begin{table}[!htb]
\centering
\caption{Statistical tests of the Godspeed Questionnaire}
\label{GodspeedTests}
\begin{tabular}{ |c|c|c|c| } 
 \hline
 Question & Test Type & Statistic & P value \\
 \hline
 Capable & T test & -1.287 & 0.206 \\
 \hline
 Responsive & T test & -0.792 & 0.434\\
 \hline
 Interactive & T test & -0.577 & 0.568\\
 \hline
 Reliable & T test & -0.584 & 0.563\\
 \hline
 Competent & Mann-Whitney U test & 128.0 & 0.061\\
 \hline
 Knowledgeable & Mann-Whitney U test & 148.5 & 0.174\\
 \hline
 Scary & T test & 0.313 & 0.757\\
 \hline
 Strange & T test & 1.925 & 0.062\\
 \hline
 Awkward & Mann-Whitney U test & 104.5 & 0.013\\
 \hline
 Dangerous & T test & 0.369 & 0.715\\ 
 \hline
\end{tabular}
\end{table}

While all the questions related to positive words resulted on a higher average for the algorithm developed in this work and a lower average on the negative words, denoting a possibility that the current algorithm was subconsciously preferred to the original one, only one of the questions had a low enough p value to be considered of any statistical significance. 

The final question provided a clearer result with $52.6\%$ of the participants preferring the current algorithm, while only $15.8\%$ preferred the original.

The distance part of the experiment was executed using the same experiment as previously with the author of this work as a participant and without the questionnaire. This experiment was run 25 times with each algorithm and its purpose was to get a definite number to the distance between the robot and the person post-approach and to collect the HSCIs.

The results of this part were evaluated by measuring the mean ($\mu$) and standard deviation ($\sigma$) of final distance between the robot and the individual, with the approach concluded. The HSCIs were also measured to guarantee that the approach would not provoke discomfort. The non-adaptive algorithm ($\mu = 1.265$ and $\sigma = 0.0470$) showed a clear difference from the adaptive one ($\mu = 1.052$ and $\sigma = 0.0295$) These results, show a clear difference in the approach distance utilizing both algorithms. The Mann-Whitney U test reported a U statistic of 1 and a p value of $7.983*10^{-10}$, confirming the statistical significance of the difference between the two experiments.

%\begin{table}[!htb]
%\centering
%\caption{Evaluation of the post-approach distance between the robot and the human}
%\label{ApproachDistance}
%\begin{tabular}{ |c|c|c| } 
% \hline
% Algorithm & Mean $(m)$ & Standard deviation $(m)$ \\
% \hline
% Original Algorithm & 1.265 & 0.0470\\ 
% \hline
% Current Algorithm & 1.052 & 0.0295\\ 
% \hline
%\end{tabular}
%\end{table}

%These results, present in Table \ref{ApproachDistance}, show a clear difference in the approach %distance utilizing both algorithms, due to the diminished frontal personal space that is utilized for %the algorithm developed in this work. The Mann-Whitney U test reported a U statistic of 1 and a p %value of $7.983*10^{-10}$, confirming the statistical significance of the difference between the two %experiments.

The real life group experiments were conducted by asking volunteer participants to stand in specific positions, and wait for the robot to approach them. The intent of this experiment was solely to test the changes made to the approach pose estimation algorithm. Experiments were run with groups of two and three people and the final approach pose in reference to the group's center noted down. The robot started each experiment behind the group so as to increase the difficulty of a proper approach. In both experiments, the original algorithm was incapable of choosing a proper approach pose, disturbing the humans. The developed algorithm was capable of identifying the approach pose chosen by the other algorithm as unsafe and approach the group through a proper approach pose. A video of some experiments with the current algorithm is available. \footnote{Real Life Group Approach Experiments - \url{https://youtu.be/tgWoHWpTFrE}}

\section{Conclusions}
\label{sec:concl}

The objective of this work is to implement a general socially reactive navigation system capable of enabling a robot to proactively approach and interact with individuals or groups of humans, while taking into account the comfort of the humans. The system is capable of adapting personal and group spaces taking into account a group's arrangement and the velocity of the humans, as well as making minor adaptations as necessary to enable a robot to approach and interact with a human. The approach pose estimation can also take into account the dynamic nature of a human and determine whether a given approach pose is valid for a specific robot.

The group space adaptations were initially tested through the use of static datasets. The experiments demonstrated an increased performance comparatively to the original algorithm, but also demonstrated that the implemented adaptations depend on the size and form of a group. The experiments show that the algorithm is capable of enabling a robot to approach groups it would have otherwise been incapable of.

The socially reactive navigation system was then implemented in ROS, via use of the navigation package, and the adaptive spaces implemented utilizing the ROS costmap.

Experiments for static environments were done in both simulated and real life environments, and demonstrated that the algorithm could enable a robot to approach and interact with human beings in a socially acceptable manner, without running the risk of causing discomfort. The experiments for dynamic environments were limited to simulated environments, but were sufficient to make an initial demonstration of the advantages of the algorithm.

There are, nonetheless, several limitations to the algorithm. It is incapable of taking into account human-object interactions. It is also incapable of determining whether a given valid approach pose is better than another. The velocity adaptation method utilized is also simplistic and unlikely to deal with situations more complex than a linearly moving person. The algorithm has also proven to be under optimized in several aspects, such as the new costmap layer updates, and can be too slow to properly address with real dynamic situations.

% Name of your BiBTeX file
\bibliographystyle{IEEEtran}
\bibliography{./Thesis-MSc-Bibliography} % Put here your own filename
%
% The following command modifies the 'emphasis' style for bibliography entries

\vfill

\end{document}